\documentclass[reqno]{amsart}
\usepackage{mathtools}

\def\tr{{\rm Tr}}

\def\N{{\mathbb N}}

\def\R{{\mathbb R}}

\def\cC{{\mathcal C}}

\def\cL{{\mathcal L}}

\def\cN{{\mathcal N}}
\def\cP{{\mathcal P}}

\def\cone{{\mathfrak C}}
\def\diag{{\rm diag}}

\def\interior{{\rm int}}

\def\rank{{\rm rank}}

\def\CostN{\cC_{\cN}}

\def\1{{\bf 1}}

\def\argmin{{\rm argmin}}

\def\X0{X_0}

\def\eqnn{\begin{eqnarray*}}
\def\eeqnn{\end{eqnarray*}}
\def\eqn{\begin{eqnarray}}
\def\eeqn{\end{eqnarray}}

\def\prf{\begin{proof}}
\def\endprf{\end{proof}}

\theoremstyle{plain}
\newtheorem{theorem}{Theorem}[section]
\newtheorem{definition}[theorem]{Definition}
\newtheorem{proposition}[theorem]{Proposition}

\newtheorem{lemma}[theorem]{Lemma}

\numberwithin{equation}{section}

\begin{document} 

\title[Geometric Structure of Deep Learning Networks]
{Geometric structure of Deep Learning networks and construction of global $\cL^2$ minimizers}

\author{Thomas Chen}
\address[T. Chen]{Department of Mathematics, University of Texas at Austin, Austin TX 78712, USA}
\email{tc@math.utexas.edu} 
\author{Patricia Mu\~{n}oz Ewald}
\address[P. Mu\~{n}oz Ewald]{Department of Mathematics, University of Texas at Austin, Austin TX 78712, USA}
\email{ewald@utexas.edu} 

\begin{abstract}
In this paper, we explicitly determine local and global minimizers of the $\cL^2$ cost function in underparametrized Deep Learning (DL) networks. Our main goal is to obtain a rigorous mathematical understanding of their geometric structure and properties; we accomplish this by a direct construction, without invoking the gradient descent flow anywhere in this work. We specifically consider $L$ hidden layers, a ReLU ramp activation function, an $\cL^2$ Schatten class (or Hilbert-Schmidt) cost function, input and output spaces $\R^Q$ with equal dimension $Q\geq1$, and hidden layers also defined on $\R^{Q}$; the training inputs are assumed to be sufficiently clustered. The training input size $N$ can be arbitrarily large - thus, we are considering the underparametrized regime. More general settings are left to future work. We construct an explicit family of minimizers for the global minimum of the cost function in the case $L\geq Q$, which we show to be degenerate. Moreover, we determine a set of $2^Q-1$ distinct degenerate local minima of the cost function. In the context presented here, the concatenation of hidden layers of the DL network is reinterpreted as the recursive application of a {\em truncation map} which "curates" the training inputs by minimizing their noise to signal ratio. 
\end{abstract}

\maketitle

\section{Introduction}

The study of Deep Learning (DL) networks lies at the center of extensive research efforts across a wide range of disciplines, and their applications in science and technology have been groundbreaking. Nevertheless, the fundamental conceptual reasons underlying their functioning have remained insufficiently well understood, and continue to be the subject of intense investigation, \cite{arora-ICM-1,grokut-1}.  The commonly employed approach to the minimization of the cost (or loss) function in DL networks invokes its associated gradient descent flow in the space of parameters (weights and biases). While extremely powerful numerical algorithms have been developed to find satisfactory approximations to the cost minimum in various practical situations, a clear mathematical understanding of the structure of minimizers remains largely elusive. 

A main goal in the current paper is to explicitly determine local and global minimizers of the cost function in underparametrized DL networks, in order to obtain a rigorous mathematical understanding of their geometric structure and properties; we accomplish this by a direct construction, without invoking the gradient descent flow.
More precisely, we analyze the geometric structure of DL networks in the context of supervised learning, with $L>1$ hidden layers, an $\cL^2$ cost function and ReLU activation function, in continuation of our recent analysis of shallow neural networks in \cite{cheewa-1}  (i.e., the case of one hidden layer, $L=1$).
%We first recall the key results from \cite{cheewa-1} that are relevant for the work at hand.  
We focus on the special case in which the input and output spaces, as well as the spaces associated to the $L>1$ hidden layers, all have equal dimension $Q$. 
The training input size $N$ can be arbitrarily large - thus, we are considering the underparametrized regime. Moreover, we assume that the training data is sufficiently clustered. This setting will allow us to illuminate the key principles of this problem most succinctly; an analysis of more general situations (varying dimensions of the spaces determining the hidden layers instead of fixed $Q$, and less structured training inputs) is left for future work. %we derive an exact degenerate global minimum of the cost function when $L\geq Q$. 

We show that each hidden layer in a DL network acts by way of a {\em truncation map} (which is a reinterpretation of the usual map defining the layer of a DL network) which reduces one of the clusters of training inputs to a point, given specific weights and biases that are explicitly constructible. For $L\geq Q$, the concatenation of hidden layers is equivalent to recursive applications of $L$ truncation maps, whereby the $Q$ clusters of training inputs are reduced to $Q$ disjoint points; this allows us to explicitly determine a family of global minimizers of the $\cL^2$ cost in the last layer.
Moreover, we explicitly determine a set of $2^Q-1$ distinct degenerate local minima of the cost function by a variation of the above construction. 
To our knowledge, these are the first rigorous results of this type. 
For some thematically related background, see for instance \cite{achmalger,hanrol,grokut-1,lcbh,manvanzde,nonreeste} and the references therein.
The fact that the hidden layers in a DL network can be reinterpreted as a nested sequence of truncation maps exhibits an intrinsic structure that is reminiscent of renormalization maps in the renormalization group analysis of quantum field theory; see for instance \cite{bcfs-1}. 

In this paper, we will focus on the geometric structure of the system in discussion using standard mathematical terminology, and will make minimal use of specialized nomenclatures specific to the neural networks literature. We will here not focus on the development of new algorithms or application scenarios; our goal in this paper is to obtain a clear, mathematically rigorous understanding of the geometric structure of cost minimizers.

\section{Definition of the Model and Main Results}
\label{sec-DL-def-1}

In this section, we introduce a Deep Learning network with $\cL^2$ cost function.  
Let $Q\in\N$, $Q\geq2$, denote the number of distinct output values. We define the output matrix
\eqn
	Y:=[y_1,\dots,y_Q] \in \R^{Q\times Q}
\eeqn 
where $y_j\in\R^{Q}$ is the $j$-th output vector.
We assume that the family $\{y_j\}$ is linearly independent, so that $Y$ is invertible.

We will first introduce the general setting for a Deep Learning network with input space of dimension $M$ and hidden layers spaces of dimension $M_\ell$ with  $\ell=1,\dots,L$. Subsequently, we will restrict our analysis to the special case $M=M_\ell=Q$. We leave a discussion of the general case for future work.

\subsection{Training inputs}
Let $\R^M$ denote the input space with the orthonormal basis $e_j=(0,\dots,0,1,0,\dots,0)^T\in\R^M$, $j=1,\dots,M$.

For $j\in\{1,\dots,Q\}$, and $N_j\in\N$, 
\eqn\label{eq-x0ji-def-1}
	x_{0,j,i}\in\R^M
	\;\;,\;\;
	i\in\{1,\dots,N_j\}
\eeqn 
denotes the $i$-th training input vector corresponding to the $j$-th output vector $y_j$. 
%We assume that all input vectors are contained in the positive sector, $\R_+^M$. 
We define the matrix of all training inputs belonging to $y_j$ by
\eqn 
	X_{0,j} := [x_{0,j,1} \cdots x_{0,j,i} \cdots x_{0,j,N_j} ] \,.
\eeqn  
Correspondence of a training input vector to the same output $y_j$ defines an equivalence relation, that is, for each $j\in\{1,\dots,Q\}$ we have $x_{0,j,i}\sim x_{0,j,i'}$ for any $i,i'\in\{1,\dots,N_j\}$. Accordingly, $X_{0,j}$ labels the equivalence class of all inputs belonging to the $j$-th output $y_j$.

Then, we define the  matrix of training inputs
\eqn 
	\X0 := [X_{0,1}\cdots X_{0,j}\cdots X_{0,Q}] \in \R^{M\times N}
\eeqn 
where $N:=\sum_{j=1}^Q N_j$.

We define the average of all training input vectors belonging to the output $y_j$,
\eqn 
	\overline{x_{0,j}}:= \frac{1}{N_j}\sum_{i=1}^{N_j} x_{0,j,i} \in \R^M
\eeqn
for $j=1,\dots,Q$,
and 
\eqn
	\Delta x_{0,j,i}:= x_{0,j,i}-\overline{x_{0,j}} \,.
\eeqn 
We denote by
\eqn	
	\overline{x} := \frac1Q\sum_{j=1}^Q %\frac{N_j}{N}
	\overline{x_{0,j}} 
\eeqn
the overall average of training inputs.
Moreover, we define
\eqn 
	\overline{X_{0,j}} := [\overline{x_{0,j}}\cdots \overline{x_{0,j}}] \in \R^{M\times N_j}
\eeqn 
and 
\eqn\label{eq-ovlnX0-def-1}
	\overline{X_0} := [\overline{X_{0,1}}\cdots \overline{X_{0,Q}}] \in \R^{M\times N} \,.
\eeqn 
We define
\eqn 
	\Delta X_{0,j} := [\Delta x_{0,j,1} \cdots \Delta x_{0,j,i} \cdots \Delta x_{0,j,N_j} ]
	\in\R^{M\times N_j} \,,
\eeqn 
and 
\eqn 
	\Delta\X0 := [\Delta X_{0,1}\cdots \Delta X_{0,j}\cdots \Delta X_{0,Q}] \in \R^{M\times N} \,,
\eeqn
where we assume that
\eqn\label{eq-DeltX0-neq0-1-2}
	\Delta X_{0,j}\neq0 \;\;{\rm for \; all} \; j=1,\dots,Q\,.
\eeqn 
Then, 
\eqn 
	X_{0} = \overline{X_0} + \Delta X_{0} \,,
\eeqn 
and we assume
\eqn\label{eq-delta-def-1-0}
	\delta := \sup_{i,j}|\Delta x_{0,j,i}| 
	< c_0 \min_j |\overline{x_{0,j}}-\overline{x}| 
	\,,
\eeqn
for a sufficiently small constant $0<c_0<\frac14$. 

Moreover, we define 
\eqn
	\overline{X_0^{red} }:= [\overline{x_{0,1}}\cdots \overline{x_{0,Q}}] \in \R^{M\times Q}
\eeqn 
where the superscript indicates that is the column-wise reduction of the matrix $\overline{X_0}$.

\subsection{Definition of the DL network} 
We define a  DL network with $L>1$ hidden layers. Let $M_\ell\in \N$, $\ell=1,\dots,L+1$. The first hidden layer is defined by
\eqn 
	X^{(1)} := \sigma(W_1 X_0 + B_1 ) 
	\;\; \in \R^{M_1\times N}\,,
\eeqn 
and letting 
\eqn
	X^{(0)}:=X_0 \,
\eeqn 
the $\ell$-th hidden layer is recursively determined by
\eqn\label{eq-Xell-def-rec-1-0}
	X^{(\ell)} := \sigma(W_\ell X^{(\ell-1)} + B_\ell ) 
	\;\; \in \R^{M_\ell\times N} \,,
\eeqn 
for $\ell=1,\dots,L$, where $\sigma$ denotes the activation function.
The terminal layer is defined by
\eqn 
	X^{(L+1)} := W_{L+1} X^{(L)} + B_{L+1}
	\;\; \in \R^{M_{L+1}\times N} \,,
\eeqn
where we do not include an activation function.

Here, we have introduced weight matrices 
\eqn 
	W_\ell\in\R^{M_{\ell}\times M_{\ell-1}} \,,
\eeqn  
and biases 
\eqn 
	B_\ell = b_\ell \; u_{N}^T \,,
\eeqn 
where 
\eqn 
	b_\ell\in\R^{M_\ell} \,,
\eeqn 
and
\eqn\label{eq-uM-def-1-0}
	u_{N} := (1,1,\dots,1)^T \;\; \in \R^{N} \,.
\eeqn  
We will consider the ReLU activation function, defined by
\eqn
	\sigma:\R^{M\times M'}&\rightarrow&\R_+^{M\times M'}
	\nonumber\\
	A=[a_{ij}] &\mapsto& [(a_{ij})_+]
\eeqn 
for arbitrary $M,M'\in\N$, where 
\eqn\label{eq-rampfct=def-1}
	(a)_+:=\max\{0,a\}
\eeqn 
is the standard ramp function. 

In this paper, we will study the special case 
\eqn\label{eq-dim-assumpt-1-0}
	M=M_\ell=Q 
	\;\;
	{\rm for \; all} \; \ell=1,\dots,L+1\,,
\eeqn 
and assume that the weight matrices
\eqn 
	W_\ell \in GL(Q)
\eeqn 
are invertible.
A generalization of our analysis to $M_\ell>Q$ is left to future work.

Each matrix $X^{(\ell)}$ then has the form  
\eqn 
	X^{(\ell)} = [X^{(\ell)}_1\cdots X^{(\ell)}_Q] \;\;\in\R^{Q\times N}
\eeqn 	
where
\eqn 
	X^{(\ell)}_j = [x^{(\ell)}_{j,i}]_{i=1}^{N_j}\;\;\in\R^{Q\times N_j}
\eeqn 
for $j=1,\dots,Q$. Notably, $N_j$ does not depend on $\ell$.

We will refer to $\R_+^Q\subset \R^Q$ as the {\em positive sector} in $\R^Q$. Clearly, on the positive sector, the activation function $\sigma$ acts as the identity, $\sigma(x)=x$ for all $x\in\R_+^Q$.
 
We define the extension of the matrix of outputs, $Y^{ext}$, by
\eqn 
	Y^{ext}:= [Y_1 \cdots Y_Q] \in \R^{Q\times N}
\eeqn 
where 
\eqn
	Y_j := [y_j \cdots y_j] \in \R^{Q\times N_j}
\eeqn
with $N_j$ copies of the same output column vector $y_j$.
Clearly, $Y^{ext}$ has full rank $Q$.

Let $\cN\in\R^{N\times N}$ be the block diagonal matrix given by
\eqn
	\cN:=\diag(N_j \1_{N_j\times N_j} \, | \, j=1,\dots,Q) \,.
\eeqn 
We introduce the inner product on $\R^{Q\times N}$
\eqn	
	(A,B)_{\cL^2_\cN}:=\tr(A\cN^{-1}B^T)
\eeqn
and 
\eqn	
	\|A\|_{\cL^2_\cN}:=\sqrt{(A,A)_{\cL^2_\cN}} \,.
\eeqn 
We define the weighted cost function
\eqn\label{eq-CostN-def-1-0-0}
	\CostN[(W_i,b_i)_{i=1}^{L+1}] := \|X^{(L+1)}-Y^{ext}\|_{\cL^2_\cN} \,.
\eeqn 
This is equivalent to 
\eqn\label{eq-CostN-def-1}
	\CostN[(W_i,b_i)_{i=1}^{L+1}]]= \sqrt{\sum_{j=1}^Q \frac{1}{N_j}\sum_{i=1}^{N_j}
	|x_{j,i}^{(L+1)}-y_j|^2_{\R^Q} } \,.
\eeqn 
The weights $\frac 1{N_j}$ ensure that the cost function is not biased by variations of training input sample sizes belonging to $y_j$.  

\begin{definition}
Given $W\in GL(Q)$, $b\in\R^Q$, and $B=b\,u_N^T$, we define the {\em truncation map}
\eqn 
	\tau_{W,b} \;:\; \R^{Q\times N} &\rightarrow& \R^{Q\times N}
	\nonumber\\
	X &\mapsto& 
	W^{-1} ( \sigma(W X + B) - B) \,,
\eeqn 
that is, $\tau_{W,b} = a_{W,b}^{-1} \circ\sigma\circ a_{W,b} $ under the affine map $a_{W,b} :X\mapsto WX+B$.
 	
We say that $\tau_{W,b}$ is rank preserving with respect to $X$ if both
\eqn
	\rank(\tau_{W,b}(X))&=&\rank(X)
	\nonumber\\
	\rank(\overline{\tau_{W,b}(X)})&=&\rank(\overline{X})
\eeqn 
hold, and that it is rank reducing otherwise.	
\end{definition}

In the following proposition, we prove the main properties of the truncation map important for the work at hand.

\begin{proposition}
\label{prop-Xl-tau-rec-1-0}
Let $X^{(\ell)}\in\R^{Q\times N}$ be given as in \eqref{eq-Xell-def-rec-1-0}. Then, recursively,
\eqn\label{eq-Xell-tau-rec-1-0}
	X^{(\ell)} 
	\; = \; W_\ell \; \tau_{W_\ell,b_\ell}(X^{(\ell-1)})  + B_\ell  
\eeqn 
for $\ell=1,\dots,L$.

In particular,
\eqn\label{eq-Xell-tau-rec-1-1}
	X^{(\ell)} 
	\; = \;  W^{(\ell)}  \; 
	\tau_{\underline{ W }^{(\ell)} ,\underline{b}^{(\ell)}} (X^{(0)}) 
	+ B^{(\ell)} 
\eeqn 
where we define
\eqn 
	\underline{ W }^{(\ell)}  &:=& (W^{(1)},\dots,W^{(\ell)})  
	\nonumber\\
	\underline{b}^{(\ell)} &:=& (b^{(1)},\dots,b^{(\ell)})  
\eeqn 
and
\eqn\label{eq-tau-uWub-def-1-0}
	\tau_{\underline{W}^{(\ell)} ,\underline{b}^{(\ell)}} 
	(X_0)
	&:=&
	\tau_{W^{(\ell)} ,b^{(\ell)}} 
	(\tau_{W^{(\ell-1)} ,b^{(\ell-1)}} (\cdots
	\tau_{W^{(2)},b^{(2)}}(\tau_{W^{(1)},b^{(1)}}(X_0) )
	\cdots ) )
	\nonumber\\
	&=&
	\tau_{W^{(\ell)} ,b^{(\ell)}} 
	(\tau_{\underline{W}^{(\ell-1)} ,\underline{b}^{(\ell-1)}} (X_0) )
\eeqn 
with
\eqn\label{eq-Well-bell-def-1-0}
 	W^{(\ell)}  &:=& W_\ell W_{\ell-1}\cdots W_1
	\nonumber\\
	b^{(\ell)} &:=& 
	\left\{
	\begin{array}{cl}W_\ell\cdots W_2 b_1+
	W_\ell\cdots W_3 b_2+\cdots + W_{\ell}b_{\ell-1}+b_\ell
	& {\rm if}\;\ell\geq2\\
	b_1&{\rm if }\;\ell=1 \,.
	\end{array}
	\right.
\eeqn  
and 
\eqn 
	B^{(\ell)}= b^{(\ell)} \; u_N^T \,.
\eeqn 
Moreover, $\sigma(X^{(\ell)})=X^{(\ell)}$ holds for every $\ell=1,\dots,L$.
\end{proposition}

The original weights and biases $W_\ell, b_\ell$ in the definition of the hidden layers \eqref{eq-Xell-def-rec-1-0} are recursively obtained from
\eqn
	W_\ell &=& W^{(\ell)}(W^{(\ell-1)})^{-1}
	\nonumber\\
	b_\ell &=&b^{(\ell)}-W_{\ell} \, b^{(\ell-1)}\,,
\eeqn 
through \eqref{eq-Well-bell-def-1-0}, for $\ell=1,\dots,L$, with $b^{(0)}\equiv0$ and $W^{(0)}\equiv\1_{Q\times Q}$.

The key point of this proposition is the fact that the vector $X^{(\ell)}$ associated to the $\ell$-th hidden layer is obtained from applying $\ell$ nested truncations to the original training input matrix $X^{(0)}=X_0$, according to \eqref{eq-tau-uWub-def-1-0}. Notably, as $\sigma$ does not appear explicitly in \eqref{eq-Xell-tau-rec-1-0}, other than in the definition of the truncation map, we observe that the effect of the hidden layers is to "curate" $X_0$ in such a way that $\sigma$ acts as the identity in each hidden layer.

Accordingly, we can define the matrices 
\eqn 
	\overline{\tau_{\underline{ W }^{(\ell)} ,\underline{b}^{(\ell)}} (X_0) }
	\;\;,\;
	\overline{(\tau_{\underline{ W }^{(\ell)} ,\underline{b}^{(\ell)}} (X_0) )^{red}}
	\;\;,\;
	\Delta (\tau_{\underline{ W }^{(\ell)} ,\underline{b}^{(\ell)}} (X_0) )
\eeqn 
for $\ell=1,\dots,L$, in analogy to $\overline{X_0}, \overline{X_0^{red}}, \Delta X_0$.

Moreover, we introduce
\eqn
	\CostN^\tau[\underline{W}^{(L)},W_{L+1},\underline{b}^{(L)},b_{L+1}]
	\equiv \CostN[(W_i,b_i)_{i=1}^{L+1}]
\eeqn 
for notational convenience.

Next, we describe the condition on the geometry of training inputs that will be used for our main theorem. 
Because the vectors $\overline{x_{0,j}}\in\R^Q$, $j=1,\dots,Q$, are linearly independent, they are the extreme points of a convex polytope (or simplex)
\eqn\label{eq-Gamma-def-1-0}
	\Gamma_{\overline{X_0^{red}}} := \Big\{x\in\R^Q \,\Big| \, x = \sum_{j=1}^Q \kappa_j \overline{x_{0,j}}  
	\;\;,\;\kappa_j\geq0\;,\;\sum_{j=1}^Q\kappa_j=1\Big\} \,.
\eeqn 
We assume that they satisfy the following quantitative properties.
Clearly, $\overline{x}$ lies in the interior of $\Gamma_{\overline{X_0^{red}}} $, as it corresponds to $0<\kappa_j=\frac1Q<1$ in \eqref{eq-Gamma-def-1-0}.
For $j=1,\dots,Q$, let 
\eqn\label{eq-fj-def-1-0}
	f_j:= -\frac{\overline{x_{0,j}}-\overline{x}}{|\overline{x_{0,j}}-\overline{x}|}
\eeqn 
be the unit vector pointing from $\overline{x_{0,j}}$ towards $\overline{x}$.
Let 
\eqn 
	\cone_\theta[f_j] := \Big\{x\in\R^Q \, | \, \angle(x,f_j)\leq\frac\theta2\Big\}
\eeqn 
denote the cone of opening angle $\theta>0$ centered along an axis in the direction of $f_j$. Moreover, let $\theta_{*,j}$ denote the smallest angle so that
\eqn\label{eq-cone-def-1-0-0-0}
	\overline{x_{0,j}}+\cone_{\theta_{*,j}}[f_j]\supset \bigcup_{j'\neq j}B_{4\delta}(\overline{x_{0,j'}})
	%\;\;,\;\;K>4
	\,,
\eeqn  
for $\delta$ as in \eqref{eq-delta-def-1-0}.
That is, the cone of opening angle $\theta_{*,j}$ emanating from  %$\partial B_{2\delta}(\overline{x_{0,j}})$ with 
the point $\overline{x_{0,j}}$ with axis parallel to $f_j$, contains the $4\delta$-balls centered at all the other $\overline{x_{0,j'}}$, $j'\neq j$. 
%Clearly, $\theta_{*,j}(K)$ is monotonically increasing in $K>0$. 
We make the assumption that  
\eqn\label{eq-thetaj-cond-1-0}
	\max_j \theta_{*,j}<\pi 
\eeqn 
holds. 
The condition \eqref{eq-thetaj-cond-1-0} infers that $\Gamma_{\overline{X_0^{red}}}$ remains convex under dislocations of order at least $O(\delta)$, of any of its extreme points.

\begin{theorem}
\label{thm-cC-uppbd-2}
Let $M=M_1=\dots=M_{L+1}=Q<N$ as in \eqref{eq-dim-assumpt-1-0}, and assume that the training inputs $X_0$ satisfy the cone condition \eqref{eq-cone-def-1-0-0-0} with \eqref{eq-thetaj-cond-1-0}. 

Moreover, assume $\tau_{\underline{ W }^{(L)} ,\underline{b}^{(L)}}$ to be rank preserving with respect to $X_0$, so that $\rank(\overline{X^{(L)}})=\rank(X^{(L)})=Q$, and let
\eqn\label{eq-cP-def-1-0-0}
	\cP := \cN^{-1}(X^{(L)})^T((X^{(L)}) \cN^{-1} (X^{(L)})^T)^{-1} (X^{(L)})
	\;\;
	\in\R^{N\times N} \,.
\eeqn 
Its transpose $\cP^T$ is an orthonormal projector in $\R^N$ with respect to the inner product $\langle u,v\rangle_{\cN}:=(u,\cN^{-1} v)$, onto the range of $(X^{(L)})^T$. In particular,  $(A\cP, B)_{\cL^2_\cN}=(A,B\cP)_{\cL^2_\cN}$ for all $A,B\in\R^{Q\times N}$. 
Letting $\cP^\perp=\1_{N\times N}-\cP$, the weighted cost function satisfies the upper bound
\eqn\label{eq-Thm3.2-CostN-1-0-0}
	\lefteqn{
	\min_{\underline{W}^{(L)},W_{L+1},\underline{b}^{(L)},b_{L+1}}
	\CostN^\tau[\underline{W}^{(L)},W_{L+1},\underline{b}^{(L)},b_{L+1}]
	}
	\nonumber\\
	%&\leq&
	%\min_{W_L,b_L}\CostN[\underline{W}^{(L)},\underline{b}^{(L)}]
	%\nonumber\\
	&=&
	\min_{\underline{W}^{(L)},\underline{b}^{(L)}}\|Y^{ext}\cP^{\perp}\|_{\cL^2_{\cN}} 
	\nonumber\\
	&=&
	\min_{\underline{W}^{(L)},\underline{b}^{(L)}}\big\|
 	Y|\Delta_2^{(L)}|^{\frac12}\big(1+\Delta_2^{(L)}\big)^{-\frac12}   
 	\big\|_{\cL^2}
	\nonumber\\
	&\leq&(1-C_0\delta_P^2) \; 
	\min_{\underline{W}^{(L)},\underline{b}^{(L)}}\|Y \; \Delta_1^{(L)}   \|_{\cL^2_{\cN}} 
	\,,
\eeqn 
for a constant $C_0\geq0$, where
\eqn\label{eq-Delt12-tau-def-0-1}
	\Delta_{2}^{(L)} &:=& 
	\Delta_1^{(L)} \cN^{-1} (\Delta_1^{(L)})^{T} 
	\nonumber\\
	\Delta_1^{(L)} &:=&  
	(\overline{(\tau_{\underline{ W }^{(L)} ,\underline{b}^{(L)}} (X_0) )^{red}})^{-1} 
	\Delta (\tau_{\underline{ W }^{(L)} ,\underline{b}^{(L)}} (X_0) ) \,,
\eeqn 
and where
\eqn 
	\delta_{P} := \sup_{j,i}\left|
	(\overline{(\tau_{\underline{ W }^{(L)} ,\underline{b}^{(L)}} (X_0) )^{red}})^{-1} 
	\Delta (\tau_{\underline{ W }^{(L)} ,\underline{b}^{(L)}}(x_{0,j,i}) )\right|
\eeqn
measures the signal to noise ratio of the truncated training input data.

Assume that the condition \eqref{eq-thetaj-cond-1-0} holds. 
%with the property that with $\delta$ as in \eqref{eq-delta-def-1-0}, 
%\eqn
%	\meas\left\{x\in \Gamma_{\overline{X_0^{red}}}\,\Big|\, {\rm dist}(x,\partial \Gamma_{\overline{X_0^{red}}}) > 4 \delta\right\} \, > \, 0 \,,
%\eeqn
Then, if $L\geq Q$, the global minimum 
\eqn\label{eq-CostN-globmin-1-0}
	\min_{\underline{W}^{(L)},W_{L+1},\underline{b}^{(L)},b_{L+1}}
	\CostN^\tau[\underline{W}^{(L)},W_{L+1},\underline{b}^{(L)},b_{L+1}] = 0
\eeqn 
is attained, and is degenerate.
The weights and biases $\underline W^{(L)}_*$, $\underline b^{(L)}_*$ minimizing \eqref{eq-Thm3.2-CostN-1-0-0} are recursively defined in \eqref{eq-Well-def-1-0}, \eqref{eq-bell-def-1-0}, respectively.
In particular, $\underline b^{(L)}_*=\underline b^{(L)}_*[\underline\mu]$ is parametrized by $\underline\mu\in\R^Q$ where the global minimum \eqref{eq-CostN-globmin-1-0} is attained for all
\eqn 
	\underline\mu\in(-D,-2\delta|u_Q|)^Q\subset\R^Q
\eeqn
for a constant $D>2\delta|u_Q|$ as defined in \eqref{eq-cone-D-def-1-0}.
In particular,
\eqn
	\tau_{\underline{ W }_*^{(L)} ,\underline{b}_*^{(L)}[\underline\mu]} (X_0) 
	&=& [\;\overline{x_{0,1}}[\mu_1] u_{N_1}^T \;\cdots\;
	\overline{x_{0,Q}}[\mu_Q] u_{N_Q}^T \; ]
	\nonumber\\
	&=:&\overline{X_0}[\underline\mu]
\eeqn 
with 
\eqn
	\overline{x_{0,j}}[\mu_j] := \overline{x_{0,j}}-\mu_j f_j \,,
\eeqn 
and 
$f_j\in\R^Q$, $j=1,\dots,Q$, as in \eqref{eq-fj-def-1-0}.

The matching of an arbitrary input $x\in\R^Q$ to a specific output $y_{j}$ where $j=j(x)$ is obtained from
\eqn\label{eq-j-match-1-1-0}
	j(x)&=&
	\argmin_j |W_*^{(L+1)}\tau_{\underline{ W }_*^{(L)} ,\underline{b}_*^{(L)}[\underline\mu]}(x) + b^{(L+1)}_* - y_j|
	\nonumber\\
	&=&\argmin_j \; d(\tau_{\underline{ W }_*^{(L)} ,\underline{b}_*^{(L)}[\underline\mu]}(x)
	\; , \; \overline{x_{0,j}}[\mu_j])
\eeqn 
for the metric $d:\R^Q\times \R^Q\rightarrow\R_+$ on the input space, defined by  
\eqn
	d(x,x') \; := \;  \big|Y(\overline{X_0^{red}}[\underline\mu])^{-1}
	(x-x')\big|
\eeqn
where
$\overline{X_0^{red}}[\underline\mu]=[\;\overline{x_{0,1}}[\mu_1]  \;\cdots\;
	\overline{x_{0,Q}}[\mu_Q]   \;]\in GL(Q)$.
\end{theorem}

The weights and biases $\underline W^{(L)}_*$, $\underline b^{(L)}_*$ minimizing \eqref{eq-Thm3.2-CostN-1-0-0} in Theorem \ref{thm-cC-uppbd-2} have the following explicit form, 
\eqn\label{eq-Well-def-1-1}
	W_*^{(\ell)} =  W_* R_\ell
\eeqn 
where $W_*\in GL(Q)$, with $\|W_*\|_{op}=1$, is defined in \eqref{eq-Wstar-def-1-0}, and $R_\ell\in SO(Q)$ satisfies
\eqn 
	R_\ell f_\ell= \frac{u_Q}{|u_Q|} \,,
\eeqn 	
and   
\eqn\label{eq-bell-def-1-1}
	b_*^{(\ell)} =-W_*^{(\ell)}\overline{x_{0,\ell}}+\mu_\ell R_\ell f_\ell 
\eeqn 	
for $\ell=1,\dots,L$. The detailed construction is given in \eqref{eq-Well-def-1-0}, \eqref{eq-bell-def-1-0}.

We remark that the truncation map exhibits the scaling invariance 
\eqn 
	\tau_{\lambda W,\lambda b}(X) = \tau_{W,b}(X)
	\;\;{\rm for \; all} \; \lambda>0 \,,
\eeqn 
as can be easily checked. 
Therefore, choosing $\lambda=\frac{1}{\|W\|_{op}}$, one may make the assumption $\|W\|_{op}=1$ in the minimization process (respectively, $\|W^{(\ell)}\|_{op}=1$ in the given case), without any loss of generality, as $b$ is arbitrary. Notably, this scaling invariance is a feature of $\sigma$ being the ramp function.

We note that \eqref{eq-j-match-1-1-0} expresses that the DL network trained with the minimizing weights and biases naturally endows the input space $\R^Q$ with the metric
\eqn
	d(x,x') = \big|Y(\overline{X_0^{red}}[\underline\mu])^{-1}
	(x-x')\big| \,.
\eeqn 
Thus, matching an input $x\in\R^Q$ to an output $y_j$ is equivalent to finding the vector $\overline{x_{0,j}}[\mu_j]$ nearest to the truncation of $x$, given by $\tau_{\underline{ W }_*^{(L)} ,\underline{b}_*^{(L)}[\underline\mu]}(x)$, relative to $d$.
This is analogous to Theorem 3.3 in \cite{cheewa-1}.

We also note that the truncation map is constructed in a manner that 
\eqn
	\tau_{\underline{ W }_*^{(L)} ,\underline{b}_*^{(L)}[\underline\mu]}(x)=\overline{x_{0,j}}[\mu_j]
	\;\;{\rm for\;all}\;x\in B_\delta(\overline{x_{0,j}}) \,.
\eeqn
That is, it maps the entire ball $B_\delta(\overline{x_{0,j}})$ to the point $\overline{x_{0,j}}[\mu_j]$. Therefore, if $x$ lies in one of the $\delta$-balls $B_\delta(\overline{x_{0,j}})$ for a specific choice of $j$, then $d(x,\overline{x_{0,j}}[\mu_j])=0$.

The cone condition \eqref{eq-fj-def-1-0} and the convexity of \eqref{eq-Gamma-def-1-0} allow us to separately deform each $\Delta X_{0,j}$ to zero, using a construction recursive in $j=1,\dots,Q$, while maintaining control of the rank preserving property of the truncation map $\tau_{\underline{ W }^{(L)} ,\underline{b}^{(L)}}$ relative to $X_0$. As a result, we obtain that $\Delta_1^{(L)} =0$, or respectively, $\delta_P=0$, which implies that the cost is zero, and thus a global minimum. 

On the other hand, we may also construct a local degenerate minimum of the following form.

\begin{theorem}
\label{thm-cC-uppbd-3}
Let $M=M_1=\dots=M_{L+1}=Q<N$ as in \eqref{eq-dim-assumpt-1-0}, and $L=Q$, and assume that the training inputs $X_0$ satisfy the cone condition \eqref{eq-cone-def-1-0-0-0} with \eqref{eq-thetaj-cond-1-0}. 
Let $\underline W^{(L)}_*$, $\underline b^{(L)}_*$ denote the minimizers obtained in Theorem \ref{thm-cC-uppbd-2}, but with $\underline b^{(L)}_*=\underline b^{(L)}_*[\underline\mu]$ parametrized by 
\eqn 
	\underline\mu\in(2\delta|u_Q|,\infty)^Q\subset\R^Q \,.
\eeqn
Then, $\underline{W}^{(L)}_*,\underline{b}^{(L)}_*[\underline\mu]$ parametrize a family of truncation maps for which $X_0$ is a fixed point,
\eqn 
	\tau_{\underline{W}^{(L)}_*,\underline{b}^{(L)}_*[\underline\mu]} (X_0) = X_0 \,.
\eeqn 
Consequently, for all $\underline\mu\in(2\delta|u_Q|,\infty)^Q$,
\eqn\label{eq-Thm3.2-CostN-2-0}
	\lefteqn{
	\min_{ {W}_{L+1}, {b}_{L+1}}
	\CostN^\tau
	[\underline W^{(L)}_*,W_{L+1},\underline b^{(L)}_*[\underline\mu],b_{L+1}]
	}
	\nonumber\\
	%&\leq&
	%\min_{W_L,b_L}\CostN[\underline{W}^{(L)},\underline{b}^{(L)}]
	%\nonumber\\ 
	%&=&
	%\big\|Y|\Delta_2^{(L)}|^{\frac12}\big(1+\Delta_2^{(L)}\big)^{-\frac12}   
 	%\big\|_{\cL^2}  
 	%\nonumber\\
 	&&\hspace{1cm}= \;
 	\big\|
 	Y|\Delta_2^{(0)}|^{\frac12}\big(1+\Delta_2^{(0)}\big)^{-\frac12}   
 	\big\|_{\cL^2}  
	\,
\eeqn 
is a degenerate local minimum,
where
\eqn\label{eq-Delt12-tau-def-0-2}
	\Delta_{2}^{(0)} &:=& 
	\Delta_1^{(0)} \cN^{-1} (\Delta_1^{(0)})^{T} \,,
\eeqn
and
\eqn\label{eq-Delt12-tau-def-0-3}
	\Delta_1^{(0)} &:=&  
	(\overline{X_0^{red}})^{-1} 
	\Delta X_0  \,
\eeqn 
are independent of $\underline\mu\in(2\delta|u_Q|,\infty)^Q$. 
\end{theorem}

We note that $\Delta_1^{(0)}$ has the following geometric meaning. In the definition of $\Gamma_{\overline{X_0^{red}}}$ given in \eqref{eq-Gamma-def-1-0}, $\kappa=(\kappa_1,\dots,\kappa_Q)^T\in\R^Q$ are barycentric coordinates. Any point $x\in\R^Q$ can be represented in terms of 
\eqn
	x=\sum_{i=1}^Q \kappa_i\overline{x_{0,i}}
	=[\overline{x_{0,1}}\cdots\overline{x_{0,Q}}]\kappa = \overline{X_0^{red}}\kappa \,,
\eeqn 
therefore,
\eqn 
	\kappa = (\overline{X_0^{red}})^{-1}x 
\eeqn 
are the barycentric coordinates of $x$. This means that 
\eqn
	\Delta_1^{(0)} =  
	(\overline{X_0^{red}})^{-1} 
	\Delta X_0
\eeqn 
is the representation of $\Delta X_0$ in barycentric coordinates with respect to \eqref{eq-Gamma-def-1-0}. An analogous geometric interpretation holds for \eqref{eq-Delt12-tau-def-0-1}, with respect to the convex polytope $\Gamma_{\overline{X_0^{red}}[\underline \mu]}$ obtained from $\Gamma_{\overline{X_0^{red}}}$ by "deforming" its extreme points via the truncation map $\tau_{\underline{W}^{(L)}_*,\underline{b}^{(L)}_*[\underline\mu]} $.

So far, with Theorem \ref{thm-cC-uppbd-2} and Theorem \ref{thm-cC-uppbd-3}, we have found a family of weights and biases parametrized by $\underline\mu\in\R^Q$ for which we obtain a degenerate local minimum for $\underline\mu\in(2\delta|u_Q|,\infty)^Q$, and a global degenerate minimum for $\underline\mu\in(-D,-2\delta|u_Q|)^Q$. As our proof of Theorem \ref{thm-cC-uppbd-2} shows, the component-wise transitional regime, of $\mu_\ell\in[-2\delta|u_Q|,2\delta|u_Q|]$, parametrizes a family of truncation maps that continuously deforms $\Delta X_{0,\ell}$ to zero, for $\ell=1,\dots,Q$. 

In particular, we obtain the following intermediate degenerate local minima. 

\begin{theorem}
\label{thm-cC-uppbd-4}
Let $M=M_1=\dots=M_{L+1}=Q<N$ as in \eqref{eq-dim-assumpt-1-0}, and $L=Q$, and assume that the training inputs $X_0$ satisfy the cone condition \eqref{eq-cone-def-1-0-0-0} with \eqref{eq-thetaj-cond-1-0}. 
Let $\underline W^{(L)}_*$, $\underline b^{(L)}_*$ denote the minimizers from Theorem \ref{thm-cC-uppbd-2},  where $\underline b^{(L)}_*=\underline b^{(L)}_*[\underline\mu]$ is parametrized by 
\eqn 
	\underline\mu\in (I_0\cup I_1)^Q\;\subset\R^Q \,.
\eeqn
where
\eqn 
	I_0:= (-D,-2\delta|u_Q|)
\eeqn 
and 
\eqn
	I_1:= (2\delta|u_Q|,\infty) \,.
\eeqn 
Let $\underline{s}\in\{0,1\}^Q$, and define 
\eqn\label{eq-Delt12-tau-def-0-4}
	\Delta_{2}[\underline{s}] &:=& 
	\Delta_1[\underline{s}]  \cN^{-1} (\Delta_1[\underline{s}] )^{T} \,,
\eeqn
and
\eqn\label{eq-Delt12-tau-def-0-4}
	\Delta_1[\underline{s}]  &:=&  
	(\overline{X_0^{red}})^{-1} 
	\Delta X_0 [\underline{s}]  \,
\eeqn 
where 
\eqn 
	\Delta X_0 [\underline{s}] = [\Delta X_{0,1} [\underline{s}] \cdots \Delta X_{0,j}[\underline{s}] \cdots \Delta X_{0,Q}[\underline{s}] ]
\eeqn
with
\eqn
	\Delta X_{0,j}[\underline{s}] := 
	\left\{
	\begin{array}{cl}
	0 &{\rm if \;} \mu_j\in I_{s_j} \; {\rm with \;} s_j = 0 \\
	\Delta X_{0,j} &{\rm if \;} \mu_j\in I_{s_j} \; {\rm with \;} s_j = 1 \,.
	\end{array}
	\right.
\eeqn 
for $j=1,\dots,Q$.
Then,   
\eqn  
	\tau_{\underline{ W }^{(L)} ,\underline{b}^{(L)}[\underline\mu]} (X_0) 
	= 
	[\;\cdots \; s_j X_{0,j}+(1-s_j)(\overline{X_{0,j}}-\mu_j f_j u_{N_j}^T )\;\cdots\;]
\eeqn  
and
\eqn\label{eq-Thm3.2-CostN-4-0}
	\lefteqn{
	\min_{ {W}_{L+1}, {b}_{L+1}}
	\CostN^\tau
	[\underline W^{(L)}_*,W_{L+1},\underline b^{(L)}_*[\underline\mu],b_{L+1}]
	}
	\nonumber\\
	%&\leq&
	%\min_{W_L,b_L}\CostN[\underline{W}^{(L)},\underline{b}^{(L)}]
	%\nonumber\\ 
	%&=&
	%\big\|Y|\Delta_2^{(L)}|^{\frac12}\big(1+\Delta_2^{(L)}\big)^{-\frac12}   
 	%\big\|_{\cL^2}  
 	%\nonumber\\
 	&&\hspace{1cm}= \;
 	\big\|
 	Y|\Delta_2[\underline{s}]|^{\frac12}\big(1+\Delta_2[\underline{s}]\big)^{-\frac12}   
 	\big\|_{\cL^2}  
	\,
\eeqn 
is a degenerate local minimum for every $\underline{s}\in\{0,1\}^Q$. 
\end{theorem}

This follows immediately from the proofs of Theorems \ref{thm-cC-uppbd-2} and \ref{thm-cC-uppbd-3}. In particular, we note that for every $\underline{s}\in\{0,1\}^Q$, the second line in \eqref{eq-Thm3.2-CostN-4-0} is independent of $\underline\mu\in \prod_{j=1}^Q I_{s_j}$. The degenerate global minimum in Theorem \ref{thm-cC-uppbd-2} corresponds to $s_j=0$ for all $j=1,\dots,Q$, and the degenerate local minimum in Theorem \ref{thm-cC-uppbd-3} corresponds to $s_j=1$ for all $j=1,\dots,Q$.

Through Theorem \ref{thm-cC-uppbd-4}, we have constructed $2^Q-1$ degenerate local minima of the cost function which are not global minima, corresponding to $\mu_j\in I_0$ or $\mu_j\in I_1$, for $j=1,\dots,Q$, excluding the case $\mu_j\in I_0$  $\forall j$. Our analysis does not provide any information about the existence or inexistence of further local minima.

\section{Proof of Proposition \ref{prop-Xl-tau-rec-1-0}}

To begin with, for $\ell=1,\dots,L$, we have, by definition,
\eqn 
	X^{(\ell)} 
	= \sigma(W_\ell X^{(\ell-1)}  + B_\ell )  \,.
\eeqn 
Hence,
\eqn 
	X^{(\ell)}  
	= \sigma(X^{(\ell)} )
\eeqn 
follows from $\sigma\circ \sigma = \sigma$. 
We have 
\eqn\label{eq-Xell-tau-def-1-0}
	X^{(\ell)} 
	&=& W_\ell \; W_\ell^{-1} ( \sigma(W_\ell X^{(\ell-1)}  + B_\ell ) - B_\ell) + B_\ell 
	\nonumber\\
	&=& W_\ell \; \tau_{W_\ell,b_\ell} (X^{(\ell-1)}) + B_\ell  \,,
\eeqn 
which proves \eqref{eq-Xell-tau-rec-1-0}.

Therefore, 
\eqn 
	X^{(\ell)} 
	&=& W_\ell \; \tau_{W_\ell,b_\ell} (X^{(\ell-1)}) + B_\ell 
	\nonumber\\
	&=& W_\ell \;  W_\ell^{-1} \sigma(W_\ell X^{(\ell-1)}  + B_\ell ) 
	\nonumber\\
	&=& W_\ell W_{\ell-1}\;  W_{\ell-1}^{-1}W_\ell^{-1} 
	\sigma(W_\ell (W_{\ell-1}\tau_{W_{\ell-1},b_{\ell-1}}(X^{(\ell-2)})+B_{\ell-1})  + B_\ell )
	\nonumber\\
	&=&W_\ell W_{\ell-1}\; (W_\ell W_{\ell-1})^{-1} 
	\sigma(W_\ell W_{\ell-1}\tau_{W_{\ell-1},b_{\ell-1}}(X^{(\ell-2)})+W_\ell B_{\ell-1} + B_\ell )
	\nonumber\\
	&=&W_\ell W_{\ell-1}\; \tau_{W_\ell W_{\ell-1}, W_\ell b_{\ell-1} + b_\ell  }
	(\tau_{W_{\ell-1},b_{\ell-1}}(X^{(\ell-2)}))+W_\ell B_{\ell-1} + B_\ell 
\eeqn 
where we used \eqref{eq-Xell-tau-def-1-0} to pass to the second line, and again to pass to the last line.
By iteration of this argument, we arrive at \eqref{eq-Xell-tau-rec-1-1}.
\qed

\section{Proof of Theorem \ref{thm-cC-uppbd-2}}

Clearly, we have that
\eqn\label{eq-Thm3.2-CostN-2-1}
	\lefteqn{
	\min_{\underline{W}^{(L)},W_{L+1},\underline{b}^{(L)},b_{L+1}}
	\CostN^\tau[\underline{W}^{(L)},W_{L+1},\underline{b}^{(L)},b_{L+1}]
	}
	\nonumber\\
	%&\leq&
	%\min_{W_L,b_L}\CostN[\underline{W}^{(L)},\underline{b}^{(L)}]
	%\nonumber\\
	&=&
	\min_{\underline{W}^{(L)},\underline{b}^{(L)}}
	\min_{W_{L+1},b_{L+1}}
	\|W_{L+1}X^{(L)}+B_{L+1}-Y^{ext}\|_{\cL^2_{\cN}} \,.
\eeqn 
Let
\eqn\label{eq-cP-def-1-0-1}
	\cP := \cN^{-1}(X^{(L)})^T((X^{(L)}) \cN^{-1} (X^{(L)})^T)^{-1} (X^{(L)})
	\;\;
	\in\R^{N\times N} \,,
\eeqn 
noting that $\cP^T$, with $\cP\cN^{-1}=\cN^{-1}\cP^T$, is a projector onto the span of $X_0^T$. Also, let $\cP^\perp:=\1_{N\times N} - \cP$.
It follows immediately from Theorem 3.2 and Theorem 3.5 in \cite{cheewa-1} that minimization in $W_{L+1},b_{L+1}$ yields
\eqn\label{eq-Thm3.2-CostN-1-0}
	\lefteqn{
	\min_{\underline{W}^{(L)},\underline{b}^{(L)}}
	\min_{W_{L+1},b_{L+1}}
	\|W_{L+1}X^{(L)}+B_{L+1}-Y^{ext}\|_{\cL^2_{\cN}} 
	}
	\nonumber\\
	%&\leq&
	%\min_{W_L,b_L}\CostN[\underline{W}^{(L)},\underline{b}^{(L)}]
	%\nonumber\\
	&=&
	\min_{\underline{W}^{(L)},\underline{b}^{(L)}}\|Y^{ext}\cP^{\perp}\|_{\cL^2_{\cN}} 
	\nonumber\\
	&=&
	\min_{\underline{W}^{(L)},\underline{b}^{(L)}}\big\|
 	Y|\Delta_2^{(L)}|^{\frac12}\big(1+\Delta_2^{(L)}\big)^{-\frac12}   
 	\big\|_{\cL^2}
	\nonumber\\
	&\leq&(1-C_0\delta_P^2) \; 
	\min_{\underline{W}^{(L)},\underline{b}^{(L)}}\|Y \; \Delta_1^{(L)}   \|_{\cL^2_{\cN}} 
	\,,
\eeqn 
for a constant $C_0\geq0$, where
\eqn
	\Delta_{2}^{(L)} &=& 
	\Delta_1^{(L)} \cN^{-1} (\Delta_1^{(L)})^{T} 
	\nonumber\\
	\Delta_1^{(L)} &=&  
	(\overline{(\tau_{\underline{ W }^{(L)} ,\underline{b}^{(L)}} (X_0) )^{red}})^{-1} 
	\Delta (\tau_{\underline{ W }^{(L)} ,\underline{b}^{(L)}} (X_0) ) \,,
\eeqn 
and where
\eqn 
	\delta_{P} := \sup_{j,i}\left|
	(\overline{(\tau_{\underline{ W }^{(L)} ,\underline{b}^{(L)}} (X_0) )^{red}})^{-1} 
	\Delta (\tau_{\underline{ W }^{(L)} ,\underline{b}^{(L)}}(x_{0,j,i}) )\right|
\eeqn
measures the signal to noise ratio of the truncated training input data.

Next, we prove the following Lemma.

\begin{lemma}\label{lm-sigma-b-1-0}
Let $B_\delta(0):=\{x\in\R^Q \, | \, |x|<\delta\}$, and $b:=\mu u_Q\in\R^Q$ where $\mu\in\R$, and $u_Q\in\R^Q$ as defined in \eqref{eq-uM-def-1-0}. Then, 
\eqn 
	\sigma(Wx+b) = 0
	\;\;\forall x\in B_\delta(0)
	\;\;,\;
	\forall \;\mu<-2\delta \|W\|_{op} \,.
\eeqn 
Let ${\bf 0}_{Q\times N_j}$ denote the $Q\times N_j$ matrix with all entries zero.
Then, in particular, for all $j=1,\dots,Q$,
\eqn\label{eq-sig-DeltX0j-b-0-1}
	\sigma(W\Delta X_{0,j} + B) = {\bf 0}_{Q\times N_j}
\eeqn
with $B=\mu \, u_Q u_{N_j}^T$, and any $\mu<-2\delta \|W\|_{op}$.
\end{lemma}

\prf
Clearly, 
\eqn
	Wx \in B_{\delta\|W\|_{op}}(0) \;\;\forall x\in B_\delta(0)\,,
\eeqn
and $B_{\delta\|W\|_{op}}(0)+b=B_{\delta\|W\|_{op}}(b)$. Given $b=\mu u_Q\in\R^Q$ with $\mu<-2\delta\|W\|_{op}$, all points $x\in B_{\delta\|W\|_{op}}(b)$ have purely negative coordinates. %(the distance of the point $|\mu| u_Q$ from any coordinate axis is $|\mu|\sqrt{1-\frac1Q}\geq\frac{\sqrt2}2 |\mu|>\sqrt2\delta\|W\|_{op}$ for $Q\geq2$). 
This is because any $x\in B_{\delta\|W\|_{op}}(b)$ has the coordinate representation
\eqn 
	x=(\mu+x_1',\dots,\mu+x_Q')^T \;
	\; {\rm with} \;\; |x'|<\delta\|W\|_{op} \,,
\eeqn 
and since $|x'_i|<\delta\|W\|_{op}$ for all $i=1,\dots,Q$,  
\eqn 
	\mu_i+x_i' \; < \;  -2\delta\|W\|_{op} + |x'_i| \; < \; - \delta\|W\|_{op} \,.
\eeqn 
Therefore, $\sigma$ maps any $x\in B_\delta(b)$ to zero (that is, to the origin).

By definition of $\delta$ in \eqref{eq-delta-def-1-0}, 
\eqn 
	\Delta x_{0,j,i} \in  B_\delta(0)
	\;\;,\;
	\forall j=1,\dots,Q
	\;\;,\;
	\forall i=1,\dots, N_j \,.
\eeqn 
Therefore, 
\eqn 
	\sigma(W\Delta x_{0,j,i}+b)= 0 \; \in\R^Q
\eeqn 
for all $j,i$, given $b=\mu u_Q\in\R^Q$ with $\mu<-2\delta\|W\|_{op}$. 

Hence, \eqref{eq-sig-DeltX0j-b-0-1} follows.
\endprf

As an immediate consequence, we have that for $W\in GL(Q)$,
\eqn  
	\tau_{W,b}( X_{0,j})
	&=&
	W^{-1}(\sigma(W X_{0,j}+B)-B)
	\nonumber\\
	&=& -W^{-1}b u_{N_j}^T
	\nonumber\\
	&=&
	W^{-1}(\overline{x_{0,j}}-\mu u_Q ) \; u_{N_j}^T
	\nonumber\\
	&=&
	W^{-1}(\overline{X_{0,j}}-\mu u_Q u_{N_j}^T)  
\eeqn 
for 
\eqn
	b=-\overline{x_{0,j}}+\mu u_Q\in\R^Q
\eeqn 
with 
\eqn
	\mu<-2\delta\|W\|_{op}\,.
\eeqn 
That is, all training inputs $x_{0,j,i}$ are mapped to the same point $-W^{-1}b$ by the truncation map parametrized by $W$ and $b$. This, in particular, has the effect that
\eqn  
	\overline{\tau_{W,b}( X_{0,j})} =  
	\tau_{W,b}( X_{0,j}) \,,
\eeqn 
or in other words,
\eqn 
	\Delta\tau_{W,b}( X_{0,j}) = 0 \,.
\eeqn 
Notably, this can be satisfied with $\tau_{W,b}$ being rank preserving.
We will exploit this mechanism to construct a degenerate global minimum of the cost function.

Along these lines, we will now recursively determine a family of weights and biases  
\eqn 
	\underline{ W }^{(\ell)}  &:=& (W^{(1)},\dots,W^{(\ell)})  
	\nonumber\\
	\underline{b}^{(\ell)} &:=& (b^{(1)},\dots,b^{(\ell)})  
\eeqn 
in such a way that 
\eqn 
	\Delta (\tau_{\underline{ W }^{(L)} ,\underline{b}^{(L)}}(X_{0,j}) ) = 0 
\eeqn 
for all $j=1,\dots,Q$. As a consequence, we obtain that $\Delta (\tau_{\underline{ W }^{(L)} ,\underline{b}^{(L)}} (X_0) )=0$, and hence, that $\Delta_1^{(L)} =0$ and $\Delta_2^{(L)} =0$, resulting in
\eqn 
	0\leq \min_{\underline{W}^{(L)},\underline{b}^{(L)}}\big\|
 	Y|\Delta_2^{(L)}|^{\frac12}\big(1+\Delta_2^{(L)}\big)^{-\frac12}   
 	\big\|_{\cL^2}\leq 0 \,.
\eeqn 
Thereby, we obtain a degenerate global minimum of the cost function.

We now give the detailed proof, through induction in $\ell=1,\dots,Q\leq L$.

\subsection{Induction base $\ell=1$}
\label{ssec-indbase-1-0}

For $j=1,\dots,Q$, let 
\eqn 
	f_j:= -\frac{\overline{x_{0,j}}-\overline{x}}{|\overline{x_{0,j}}-\overline{x}|}
\eeqn 
be the unit vector pointing from $\overline{x_{0,j}}$ towards $\overline{x}$.
We recall that 
\eqn 
	\cone_\theta[f_j] := \Big\{x\in\R^Q \, | \, \angle(x,f_j)\leq\frac\theta2\Big\}
\eeqn 
denotes the cone of opening angle $\theta>0$ centered around an axis in the direction of $f_j$. We note that \eqref{eq-cone-def-1-0-0-0} is equivalent with the statement that $\theta_{*,j}$ is the smallest angle so that
\eqn\label{eq-cone-def-1-0-1} 
	\overline{x_{0,j}}+2\delta f_j +\cone_{\theta_{*,j}}[f_j]\supset \bigcup_{j'\neq j}B_{2\delta}(\overline{x_{0,j'}})
	%\;\;,\;\;\forall |\mu|\in [2\delta,D] 
	\,.
\eeqn 
This format of the statement will be convenient for later use.
That is, the cone of opening angle $\theta_{*,j}$ emanating from  %$\partial B_{2\delta}(\overline{x_{0,j}})$ with 
the point $\overline{x_{0,j}}+2\delta f_j$ in the interior of $\Gamma_{\overline{X_0}}$, and with axis parallel to $f_j$, contains the $2\delta$-balls centered at all the other $\overline{x_{0,j'}}$, $j'\neq j$. By assumption \eqref{eq-thetaj-cond-1-0}, we have that $\max_j \theta_{*,j}<\pi$.
Then, let
\eqn\label{eq-thet0-def-1-0}
	\theta_* := \theta_0+\max_j \theta_{*,j}
\eeqn 
for a constant $\theta_0>0$, such that
\eqn 
	\theta_*<\pi \,.
\eeqn 
Finally, we assume $D>2\delta|u_Q|$ to be the largest number so that for every $j=1,\dots,Q$,
\eqn\label{eq-cone-D-def-1-0}
	-\mu f_j + \cone_{\theta_*}[f_j] \;\; \supset \;\; \bigcup_{j'\neq j}
	B_\delta(\overline{x_{0,j'}}-\overline{x_{0,j}})
	\;\;,\;
	\forall \mu\in(-D,-2\delta|u_Q|)
\eeqn
This condition means that for each $j$, the cone $\cone_{\theta_*}[f_j]$ can be translated by $|\mu|\in[2\delta|u_Q|,D]$ in the direction of $f_j$ into the interior of $\Gamma_{\overline{X_0^{red}}}$, in such a way that it contains all of the $B_\delta(\overline{x_{0,j'}})$, $j'\neq j$, but not $B_\delta(\overline{x_{0,j}})$. We have introduced the parameter $\theta_0>0$ in \eqref{eq-thet0-def-1-0} to control the length of the interval $(-D,-2\delta|u_Q|)$, as $D$ is monotone increasing in $\theta_0$.

Next, we define
\eqn
	\theta_Q := 2\arccos\frac{\sqrt{Q-1}}{\sqrt{Q}}
\eeqn
and we note that $\cone_{\theta_Q } [u_Q]$ is the largest cone with axis aligned with $u_Q$ that is contained in the positive sector of $\R^Q$. Let $W_*\in GL(Q)$ be given by
\eqn\label{eq-Wstar-def-1-0}
	W_* = \widetilde R \;
	\diag(1,\lambda(\theta_*,\theta_Q),\lambda(\theta_*,\theta_Q),\dots,\lambda(\theta_*,\theta_Q)) \;
	\widetilde R^T
\eeqn
with
\eqn
	\lambda(\theta_*,\theta_Q):=\left\{
	\begin{array}{cc}
		\frac{\tan\frac{\theta_Q}2}{\tan\frac{\theta_*}2}\;<1&{\rm if\;}\theta_*>\theta_Q \\
		1&{\rm otherwise},
	\end{array}
	\right.
\eeqn
where $\widetilde R\in SO(Q)$ maps the $e_1$ coordinate axis to the diagonal, $\widetilde R e_1=\frac{u_Q}{|u_Q|}$.
Then, $W_*$ has the property that 
\eqn 
	W_* x  \subset \cone_{\theta_Q } [u_Q]
	\;\;\forall x\in \cone_{\theta_*}[u_Q] \,,
\eeqn 
and in particular, 
\eqn\label{eq-WuQ-uQ-1-0}
	W_* u_Q=u_Q \,.
\eeqn
That is, $W_*$ maps the cone of opening angle $\theta_*$ centered along the diagonal $u_Q\in \R^Q$ into the interior of the cone of opening angle $\theta_Q$, which is also centered along $u_Q$. This ensures that the image of $\cone_{\theta_*}[u_Q]$ under $W_*$ is contained in the positive sector of $\R^Q$, and as a consequence, $\sigma(W_* x)=W_*x$ for all $x\in \cone_{\theta_*}[u_Q]$.
Moreover, $\|W_*\|_{op}=1$, and $\|W_*^{-1}\|_{op}=\frac1{\lambda(\theta_*,\theta_Q)}$.

For $j=1$, let 
\eqn
	W_1 :=  W_* R_1
\eeqn 
with an orthogonal matrix $R_1\in SO(Q)$ satisfying
\eqn 
	R_1 f_1 = \frac{u_Q}{|u_Q|} \,,
\eeqn 	
and let 
\eqn 
	b_1 := -W_1\overline{x_{0,1}}+\mu_1 R_1 f_1 \,.
\eeqn 	
Then, we find that
\eqn\label{eq-tauW1b1-X01-1-0}
	\tau_{W_1,b_1}(X_{0,1}) 
	&=& W_1^{-1}(\sigma(W_1X_{0,1}+b_1 u_{N_1}^T)-b_1 u_{N_1}^T)
	\nonumber\\
	&=&
	W_1^{-1}(\sigma( W_* R_1 \Delta X_{0,1}+\mu_1 R_1 f_1 u_{N_1}^T)-b_1 u_{N_1}^T)
	\nonumber\\
	&=&
	W_1^{-1}(\sigma(W_* R_1 \Delta X_{0,1}+\mu_1 \frac{u_Q}{|u_Q|} u_{N_1}^T)-b_1 u_{N_1}^T)
	\nonumber\\
	&=&
	-W_1^{-1}b_1 u_{N_1}^T
	\nonumber\\
	&=&
	(\overline{x_{0,1}} - \mu_1 R_1^T W_*^{-1} R_1   f_1) u_{N_1}^T
	\nonumber\\
	&=&
	(\overline{x_{0,1}} - \mu_1   f_1) u_{N_1}^T
\eeqn 
for $\mu_1<-2\delta|u_Q|$, using that $\|W_1\|_{op}=1$, and the fact that
\eqn\label{eq-WinvuQ-uQ-1-0}
	W_*^{-1}R_1   f_1 = W_*^{-1}\frac{u_Q}{|u_Q|} = \frac{u_Q}{|u_Q|}
\eeqn 
due to \eqref{eq-WuQ-uQ-1-0}.
In particular, 
\eqn 
	\Delta \tau_{W_1,b_1}(X_{0,1}) = 0 \,,
\eeqn 
and moreover,
\eqn
	\tau_{W_1,b_1}(X_{0,1}) 
	= (\overline{x_{0,1}} - \mu_1   f_1) u_{N_1}^T 
\eeqn 
where
\eqn 
	\overline{x_{0,1}} - \mu_1   f_1
	\;\;\in\;\interior(\Gamma_{\overline{X_0^{red}}}) \,,
\eeqn 
since $\mu_1<0$. Here, $\interior(\Gamma_{\overline{X_0^{red}}})$ denotes the interior of $\Gamma_{\overline{X_0^{red}}}$.
On the other hand, for $j\neq 1$, we have that 
\eqn\label{eq-tauW1b1-X0j-1-0}
	\tau_{W_1,b_1}(X_{0,j}) 
	&=&
	W_1^{-1}(\sigma(W_1X_{0,j}+b_1 u_{N_j}^T)-b_1 u_{N_j}^T)
	\nonumber\\
	&=&
	W_1^{-1}(\sigma(W_* R_1 (X_{0,j}-\overline{x_{0,1}}u_{N_j}^T)+ \mu_1 R_1 f_1 u_{N_j}^T)-b_1 u_{N_j}^T)
	\nonumber\\
	&=&
	W_1^{-1}(\sigma(W_* R_1 (X_{0,j}-\overline{x_{0,1}}u_{N_j}^T)+ \mu_1 \frac{u_Q}{|u_Q|} u_{N_j}^T)-b_1 u_{N_j}^T)
	\nonumber\\
	&=&
	W_1^{-1}(W_* R_1 (X_{0,j}-\overline{x_{0,1}}u_{N_j}^T)+ \mu_1 \frac{u_Q}{|u_Q|} u_{N_j}^T-b_1 u_{N_j}^T)
	\nonumber\\
	&=&
	W_1^{-1}(W_1X_{0,j}+b_1 u_{N_j}^T-b_1 u_{N_j}^T)
	\nonumber\\
	&=&
	X_{0,j}
\eeqn
To pass to the fourth line, we used that
\eqn\label{eq-tauW1b1-X0j-2-0}
	\lefteqn{
	\sigma(W_* R_1 (X_{0,j}-\overline{x_{0,1}}u_{N_j}^T)+ \mu_1 \frac{u_Q}{|u_Q|} u_{N_1}^T)
	}
	\nonumber\\
	&=&
	\sigma((W_* R_1 (\overline{x_{0,j}}-\overline{x_{0,1}}) + \mu_1 \frac{u_Q}{|u_Q|} ) u_{N_1}^T + W_1\Delta X_{0,j})
	\nonumber\\
	&=&
	(W_* R_1 (\overline{x_{0,j}}-\overline{x_{0,1}}) + \mu_1 \frac{u_Q}{|u_Q|} ) u_{N_1}^T + W_1\Delta X_{0,j}
	\nonumber\\
	&=&
	W_* R_1 (X_{0,j}-\overline{x_{0,1}}u_{N_j}^T)+ \mu_1 \frac{u_Q}{|u_Q|} u_{N_1}^T
\eeqn 
This is because $(\overline{x_{0,j}}-\overline{x_{0,1}})$ lies inside the cone $\cone_{\theta_*}[f_1]$. Application of $R_1$ maps $\cone_{\theta_*}[f_1]$ to $\cone_{\theta_*}[u_Q]$, and $W_*$ maps $\cone_{\theta_*}[u_Q]$ into $\cone_{\theta_Q}[u_Q]$, so that every column vector of $W_* R_1 (X_{0,j}-\overline{x_{0,1}}u_{N_j}^T)$ lies inside $\cone_{\theta_Q}[u_Q]$. Moreover, $\sup_{j,i}|W_1\Delta x_{0,j,i}|\leq \delta$, as $\|W_1\|_{op}=\|W_*\|_{op}=1$. Therefore, 
\eqn 
	W_* R_1 (\overline{x_{0,j}}-\overline{x_{0,1}})   + W_1\Delta x_{0,j,i} \in 
	B_{\delta}(W_* R_1 (\overline{x_{0,j}}-\overline{x_{0,1}})) \,.
\eeqn 
Thus, the  $i$-th column of the matrix expression on the third line in \eqref{eq-tauW1b1-X0j-2-0} satisfies
\eqn\label{eq-tauW1b1-X0j-3-0}
	\lefteqn{
	W_* R_1 (\overline{x_{0,j}}-\overline{x_{0,1}})  + \mu_1 \frac{u_Q}{|u_Q|}  + W_1\Delta x_{0,j,i} 
	}
	\nonumber\\
	&\in& 
	\mu_1 \frac{u_Q}{|u_Q|} + B_{\delta}(W_* R_1 (\overline{x_{0,j}}-\overline{x_{0,1}}))
	\nonumber\\
	&\subset&
	\cone_{\theta_Q}[u_Q]
\eeqn 
for all $i=1,\dots,N_j$, and for $-D\leq \mu_1\leq -2\delta|u_Q|$, due to \eqref{eq-cone-D-def-1-0}. $\cone_{\theta_Q}[u_Q]$ lies inside the positive sector $\R_+^Q\subset\R^Q$, therefore $\sigma$ acts on the l.h.s. of \eqref{eq-tauW1b1-X0j-3-0} as the identity, and we may pass from the third to the fourth line in \eqref{eq-tauW1b1-X0j-1-0}.

\subsection{The induction step $\ell-1\rightarrow\ell$}
\label{ssec-indstep-1-0}

We now prove the induction step in the recursion for $\ell-1\rightarrow\ell$.

From \eqref{eq-tau-uWub-def-1-0}, we have
\eqn\label{eq-tau-uWub-def-1-1}
	\tau_{\underline{W}^{(\ell)} ,\underline{b}^{(\ell)}} 
	(X_0)
	=
	\tau_{W^{(\ell)} ,b^{(\ell)}} 
	(\tau_{\underline{W}^{(\ell-1)} ,\underline{b}^{(\ell-1)}} (X_0) ) \,.
\eeqn 
For brevity of notation, we define
\eqn 
	X_{0,j}^{(\tau, \,\ell-1)} := \tau_{\underline{W}^{(\ell-1)} ,\underline{b}^{(\ell-1)}} (X_{0,j}) 
	\;\;,\;
	j=1,\dots,Q \,,
\eeqn 
and
\eqn 
	\overline{x_{0,j}^{(\tau, \,\ell-1)} } 
	&:=&\frac{1}{N_j}
	\sum_{i=1}^{N_j} 
	\tau_{\underline{W}^{(\ell-1)} ,\underline{b}^{(\ell-1)}} (x_{0,j,i}) 
	\;\;,\;
	j=1,\dots,Q \,,
\eeqn 
as well as 
\eqn 
	\overline{x^{(\tau,\,\ell-1)} } := \frac1Q \sum_{j=1}^Q\overline{x_{0,j}^{(\tau, \,\ell-1)} } \,.
\eeqn 
Moreover,
\eqn 
	\Delta x_{0,j,i}^{(\tau, \,\ell-1)} 
	&:=&
	\tau_{\underline{W}^{(\ell-1)} ,\underline{b}^{(\ell-1)}} (x_{0,j,i})
	-  \overline{x_{0,j}^{(\tau, \,\ell-1)} } 
	\;\;,\;
	j=1,\dots,Q \,,
\eeqn 
and
\eqn 
	\Delta X_{0,j}^{(\tau, \,\ell-1)} 
	&:=&
	\tau_{\underline{W}^{(\ell-1)} ,\underline{b}^{(\ell-1)}} (X_{0,j})
	-  \overline{x_{0,j}^{(\tau, \,\ell-1)} } u_{N_j}^T
	\;\;,\;
	j=1,\dots,Q \,.
\eeqn 
We assume that $W_r,b_r$, $r=1,\dots,\ell-1$ have been determined for the following to hold.

\subsubsection{Induction hypotheses} $\;$\\
\noindent$\bullet$ For $j\leq\ell-1$, we have
\eqn\label{eq-Delt-X0jell-0-1-0}
	\Delta\tau_{\underline{W}^{(\ell-1)} ,\underline{b}^{(\ell-1)}} (X_{0,j}) =
	\Delta X_{0,j}^{(\tau, \,\ell-1)} = 0
	\,,
\eeqn 
and
\eqn 
	\overline{ X_{0,j}^{(\tau, \,\ell-1)} } 
	&=& X_{0,j}^{(\tau, \,\ell-1)}
	\nonumber\\
	&=&
	(\overline{x_{0,j}} - \mu_j   f_j) u_{N_j}^T 
\eeqn 
where
\eqn 
	\overline{x_{0,j}} - \mu_j   f_j
	  \;\;\in\;\interior(\Gamma_{\overline{X_0^{red}}})
\eeqn 
for $\mu_j\in(-D,-2\delta|u_Q|)$.

\noindent$\bullet$ For $j\geq\ell$, we have
\eqn\label{eq-X0jell-0-1-0}
	X_{0,j}^{(\tau, \,\ell-1)}
	=\tau_{\underline{W}^{(\ell-1)} ,\underline{b}^{(\ell-1)}} (X_{0,j}) = X_{0,j} 
\eeqn  
and 
\eqn\label{eq-X0jell-0-1-0}
	\Delta X_{0,j}^{(\tau, \,\ell-1)}
	=\Delta  X_{0,j} 
\eeqn  
so that
\eqn 
	\sup_{j,i}|\Delta\tau_{\underline{W}^{(\ell-1)} ,\underline{b}^{(\ell-1)}} 
	(x_{0,j,i})|
	=
	\sup_{j,i}|\Delta x_{0,j,i}|
	<\delta  \,.
\eeqn

\subsubsection{Induction step}
\label{sssec-indstep-1-0}
%We 
%define
%\eqn 
%	f_j^{(\ell)}:=-\frac{\overline{x_{0,j}^{(\ell-1)} } -\overline{x^{(\ell-1)} } }
%	{| \overline{x_{0,j}^{(\ell-1)} } -\overline{x^{(\ell-1)} } |}
%\eeqn 
%and note that for $j=\ell$, we have 
%\eqn
%	f_\ell^{(\ell)} =f_\ell \,.
%\eeqn 
We choose 
\eqn\label{eq-Well-def-1-0}
	W^{(\ell)} :=  W_* R_\ell
\eeqn 
where $R_\ell\in SO(Q)$ satisfies
\eqn 
	%R_\ell f_\ell^{(\ell)} = 
	R_\ell f_\ell= \frac{u_Q}{|u_Q|} \,,
\eeqn 	
and   
\eqn\label{eq-bell-def-1-0}
	b^{(\ell)} &:=& -W^{(\ell)}\overline{x_{0,\ell}^{(\tau, \,\ell-1)}}+\mu_\ell R_\ell f_\ell
	\nonumber\\
	&=&-W^{(\ell)}\overline{x_{0,\ell}}+\mu_\ell R_\ell f_\ell \,.
\eeqn 	
Then, we obtain the following.

\noindent{$\bullet$}
For $j<\ell$, we find 
\eqn\label{eq-tau-uWub-2-0}
	\tau_{\underline{W}^{(\ell)} ,\underline{b}^{(\ell)}} 
	(X_{0,j})
	&=&
	\tau_{W^{(\ell)} ,b^{(\ell)}} 
	(X_{0,j}^{(\tau,\,\ell-1)} )
	\nonumber\\
	&=&
	\tau_{W^{(\ell)} ,b^{(\ell)}} 
	(\overline{X_{0,j}^{(\tau, \,\ell-1)}})
	\nonumber\\
	&=&
	\overline{X_{0,j}^{(\tau, \,\ell)}}
	\\
	&=&(\overline{x_{0,j}} - \mu_j   f_j) u_{N_j}^T 
	\nonumber
\eeqn 
where
\eqn 
	\overline{x_{0,j}} - \mu_j   f_j
	\;\;\in\;\interior(\Gamma_{\overline{X_0^{red}}})
\eeqn 
for $\mu_j\in(-D,-2\delta|u_Q|)$, $j=1,\dots,\ell-1$, using the induction assumption \eqref{eq-Delt-X0jell-0-1-0}.

\noindent{$\bullet$}
For $j=\ell$, we have  
\eqn 
	\tau_{\underline{ W }^{(\ell)} ,\underline{b}^{(\ell)}}(X_{0,\ell}) 
	&=& (W^{(\ell)})^{-1}(\sigma(W^{(\ell)} X_{0,\ell }^{(\tau, \,\ell-1)}+b^{(\ell)} u_{N_\ell }^T)-b^{(\ell)} u_{N_\ell}^T)
	\nonumber\\
	&=&
	(W^{(\ell)})^{-1}(\sigma( W_* R_\ell \Delta X_{0,\ell}^{(\tau, \,\ell-1)}
	+\mu_\ell R_\ell f_\ell%^{(\ell)} 
	u_{N_\ell}^T)-b^{(\ell)} u_{N_\ell}^T)
	\nonumber\\
	&=&
	(W^{(\ell)})^{-1}(\sigma(W_* R_\ell \Delta X_{0,\ell}
	+\mu_\ell \frac{u_Q}{|u_Q|} u_{N_1}^T)-b^{(\ell)} u_{N_\ell}^T)
	\nonumber\\
	&=&
	-(W^{(\ell)})^{-1}b^{(\ell)} u_{N_\ell}^T
	\nonumber\\
	&=&
	(\overline{x_{0,\ell}} - \mu_\ell R_\ell^T W_*^{-1} R_\ell   f_\ell )
	u_{N_\ell}^T 
	\nonumber\\
	&=&
	(\overline{x_{0,\ell}} - \mu_\ell  f_\ell )
	u_{N_\ell}^T  \,,
\eeqn 
where 
\eqn
	\overline{x_{0,\ell}} - \mu_\ell  f_\ell 
	\;\;\in\;\interior(\Gamma_{\overline{X_0^{red}}})
\eeqn 
using \eqref{eq-X0jell-0-1-0}, and the same argument as for \eqref{eq-tauW1b1-X01-1-0} with $\mu_\ell<-2\delta|u_Q|$, as well as \eqref{eq-WinvuQ-uQ-1-0}.
In particular, this implies that
\eqn 
	\Delta\tau_{\underline{ W }^{(\ell)} ,\underline{b}^{(\ell)}}(X_{0,\ell}) = 0
\eeqn 
holds.

\noindent{$\bullet$}
For $j>\ell$, we %have $f_j^{(\ell)} =f_j$. We 
invoke the same argument as in as in 
\eqref{eq-tauW1b1-X0j-2-0}, using the fact that $x_{0,j,i}-\overline{x_{0,\ell}} \in B_\delta(\overline{x_{0,j}}-\overline{x_{0,\ell}})\subset\cone_{\theta_*}[f_\ell]$ for all $i=1,\dots,N_j$, to obtain
\eqn\label{eq-tauW1b1-X0j-1-1}
	\tau_{\underline{ W }^{(\ell)} ,\underline{b}^{(\ell)}}(X_{0,j}) 
	&=&
	(W^{(\ell)})^{-1}(\sigma((W^{(\ell)})X_{0,j}^{(\tau, \,\ell)}+b^{(\ell)}  u_{N_j}^T)-b^{(\ell)}u_{N_j}^T)
	\nonumber\\
	&=&
	(W^{(\ell)})^{-1}(\sigma(W_* R_\ell (X_{0,j}-\overline{x_{0,\ell}}u_{N_j}^T)+ \mu_\ell R_\ell f_\ell u_{N_j}^T)-b^{(\ell)} u_{N_j}^T)
	\nonumber\\
	&=&
	(W^{(\ell)})^{-1}(\sigma(W_* R_\ell (X_{0,j}-\overline{x_{0,\ell}}u_{N_j}^T)+ \mu_\ell \frac{u_Q}{|u_Q|} u_{N_j}^T)-b^{(\ell)} u_{N_j}^T)
	\nonumber\\
	&=&
	(W^{(\ell)})^{-1}(W_* R_\ell (X_{0,j}-\overline{x_{0,\ell}}u_{N_j}^T)+ \mu_\ell \frac{u_Q}{|u_Q|} u_{N_j}^T-b^{(\ell)} u_{N_j}^T)
	\nonumber\\
	&=&
	(W^{(\ell)})^{-1}(W^{(\ell)}X_{0,j}+b^{(\ell)} u_{N_j}^T-b^{(\ell)} u_{N_j}^T)
	\nonumber\\
	&=&
	X_{0,j}
\eeqn
for $-D<\mu_\ell<-2\delta|u_Q|$.

This completes the proof of the induction step.

\subsection{Conclusion of the proof}
Upon completion of $L=Q$ iteration steps labeled by $\ell=1,\dots,Q$, we arrive at the construction of a family of weights and biases parametrized by $\underline{\mu}:=(\mu_1,\dots,\mu_Q)\in(-D,-2\delta|u_Q|)^Q$, which we denote here by $\underline{ W }^{(L)}_*[\underline{\mu}] ,\underline{b}^{(L)}_*[\underline{\mu}]$, so that  
\eqn\label{eq-Delttau-0-1-0} 
	\Delta \tau_{\underline{ W }^{(L)}_*[\underline{\mu}] ,\underline{b}^{(L)}_*[\underline{\mu}]}(X_{0}) = 0 \,.
\eeqn 
If $L>Q$, application of $L-Q$ additional truncation maps labeled by $\ell=Q+1,\dots,L$ does not alter \eqref{eq-Delttau-0-1-0}; the corresponding weights and biases $W^{(\ell)},b^{(\ell)}$ with $\ell=Q+1,\dots,L$ can be picked arbitrarily, under the condition that $W^{(\ell)}$, as well as
\eqn
	\overline{(\tau_{\underline{ W }^{(L)}_*[\underline{\mu}] ,
	\underline{b}^{(L)}_*[\underline{\mu}]} (X_0) )^{red}}\in\R^{Q\times Q} \,,
\eeqn 
remain invertible. 
	
Therefore, we conclude that for $L\geq Q$,
\eqn
	\Delta_{1,*}^{(L)}[\underline{\mu}] :=  
	(\overline{(\tau_{\underline{ W }^{(L)}_*[\underline{\mu}] ,
	\underline{b}^{(L)}_*[\underline{\mu}]} (X_0) )^{red}})^{-1} 
	\Delta (\tau_{\underline{ W }^{(L)}_*[\underline{\mu}] ,
	\underline{b}^{(L)}_*[\underline{\mu}]} (X_0) )
	= 0 \,,
\eeqn 
so that 
\eqn
	\Delta_{2,*}^{(L)}[\underline{\mu}] :=  
	\Delta_{1,*}^{(L)}[\underline{\mu}] \cN^{-1} (\Delta_{1,*}^{(L)})^{T}[\underline{\mu}] 
	= 0 \,,
\eeqn 
as a consequence of which we find from \eqref{eq-Thm3.2-CostN-2-0} and \eqref{eq-Thm3.2-CostN-1-0}
\eqn\label{eq-Thm3.2-CostN-1-2}
	\lefteqn{
	\min_{\underline{W}^{(L)},W_{L+1},\underline{b}^{(L)},b_{L+1}}
	\CostN^\tau[\underline{W}^{(L)},W_{L+1},\underline{b}^{(L)},b_{L+1}] 
	}
	\nonumber\\ 
	&\leq&
	\min_{W_{L+1},b_{L+1}}
	\CostN^\tau[\underline{ W }^{(L)}_*[\underline{\mu}],W_{L+1} ,\underline{b}^{(L)}_*[\underline{\mu}],b_{L+1}]
	\nonumber\\
	&=&
	\big\|
 	Y|\Delta_{2,*}^{(L)}[\underline{\mu}]|^{\frac12}\big(1+\Delta_{2,*}^{(L)}[\underline{\mu}]\big)^{-\frac12}   
 	\big\|_{\cL^2}
 	\nonumber\\
	&\leq& 
	\|Y \; \Delta_{1,*}^{(L)}[\underline{\mu}]   \|_{\cL^2_{\cN}} 
	\nonumber\\
	&=&0
	\,.
\eeqn 
Because the cost function is non-negative, this is an explicit global minimum. Moreover, because it is zero for the family of weights and biases $\underline{ W }^{(L)}_* ,\underline{b}^{(L)}_*[\underline{\mu}]$ parametrized by $\underline{\mu}\in(-D,-2\delta|u_Q|)^Q$, the minimum of the cost function is degenerate.

From our construction follows that
\eqn\label{eq-tauWb-X0-trunc-1-0-0}
	\tau_{\underline{ W }_*^{(L)} ,\underline{b}_*^{(L)}[\underline\mu]} (X_0) 
	&=& [\;\overline{x_{0,1}}[\mu_1] u_{N_1}^T \;\cdots\;
	\overline{x_{0,Q}}[\mu_Q] u_{N_Q}^T \; ]
	\nonumber\\
	&=:&\overline{X_0}[\underline\mu]
	\;\;\in\;\R^{Q\times N}
\eeqn 
with 
\eqn
	\overline{x_{0,j}}[\mu_j] := \overline{x_{0,j}}-\mu_j f_j \,,
\eeqn 
and 
$f_j\in\R^Q$, $j=1,\dots,Q$, as in \eqref{eq-fj-def-1-0}.

Finally, we prove \eqref{eq-j-match-1-1-0}.
First of all, matching an arbitrary input $x\in\R^Q$ to an output $y_{j}$ for a specific $j=j(x)$ is obtained from
\eqn\label{eq-j-match-1-1-2}
	j(x)&=&
	\argmin_j |W_*^{(L+1)}\tau_{\underline{ W }_*^{(L)} ,\underline{b}_*^{(L)}[\underline\mu]}(x) + b^{(L+1)}_* - y_j| \,.
\eeqn
From our construction, we have that
\eqn\label{eq-CostN-Y-1-1-0}
	\lefteqn{
	\|W_*^{(L+1)}\tau_{\underline{ W }_*^{(L)} ,\underline{b}_*^{(L)}[\underline\mu]}(X_0) + B^{(L+1)}_* - Y^{ext}\|_{\cL^2_\cN}
	}
	\nonumber\\
	&=&
	\|W_{L+1}^*W_*^{(L)}\tau_{\underline{ W }_*^{(L)} ,\underline{b}_*^{(L)}[\underline\mu]}(X_0) - Y^{ext}\|_{\cL^2_\cN}
	\nonumber\\
	&=&
	\|W_{L+1}^*W_*^{(L)}\tau_{\underline{ W }_*^{(L)} ,\underline{b}_*^{(L)}[\underline\mu]}(X_0) - Y^{ext}\|_{\cL^2_\cN}
	\nonumber\\
	&=&
	\|W_{L+1}^*W_*^{(L)}\overline{X_0}[\underline\mu] - Y^{ext}
	\|_{\cL^2_\cN}
\eeqn 
because $b^{(L+1)}_* =0$ (for more details, see the proof of Theorem 3.2 in \cite{cheewa-1}), and where we used \eqref{eq-tauWb-X0-trunc-1-0-0}.
The matrix $W_{L+1}^*\in GL(Q)$ minimizes the cost \eqref{eq-CostN-Y-1-1-0}, which attains the global minimum value of zero for
\eqn\label{eq-WstarL-Y-1-1-0}
	W_{L+1}^* W_*^{(L)}\overline{X_0^{red}}[\underline\mu]= Y \,,
\eeqn 
(which is equivalent to $W_{L+1}^*W_*^{(L)}\overline{X_0}[\underline\mu] = Y^{ext}$) so that
\eqn 
	W_{L+1}^* = Y (\overline{X_0^{red}}[\underline\mu])^{-1 } (W_*^{(L)})^{-1}\,,
\eeqn
where $\overline{X_0^{red}}[\underline\mu]=[\;\overline{x_{0,1}}[\mu_1]  \;\cdots\;
	\overline{x_{0,Q}}[\mu_Q]   \;]\in GL(Q)$.
Inserting this into the r.h.s. of \eqref{eq-j-match-1-1-2} yields that
\eqn 
	j(x)&=&\argmin_j \big|Y(\overline{X_0^{red}}[\underline\mu])^{-1}
	(\tau_{\underline{ W }_*^{(L)} ,\underline{b}_*^{(L)}[\underline\mu]}(x)
	-\overline{x_{0,j}}[\mu_j])\big|
\eeqn 
because the $j$-th column vector of \eqref{eq-WstarL-Y-1-1-0} reads
\eqn
	y_j = Y(\overline{X_0^{red}}[\underline\mu])^{-1} \overline{x_{0,j}}[\mu_j] \,.
\eeqn 
This concludes the proof of Theorem \ref{thm-cC-uppbd-2}.
\qed
 
\section{Proof of Theorem \ref{thm-cC-uppbd-3}}

Complementary to Lemma \ref{lm-sigma-b-1-0}, we have the following Lemma.

\begin{lemma}\label{lm-sigma-b-2-0}
Let $B_\delta(0):=\{x\in\R^Q \, | \, |x|<\delta\}$, and $b:=\mu u_Q\in\R^Q$ where $\mu\in\R$, and $u_Q\in\R^Q$ as defined in \eqref{eq-uM-def-1-0}. Then, 
\eqn 
	\sigma(Wx+b) = Wx+b
		\;\;\forall x\in B_\delta(0)
	\;\;,\;
	\forall \;\mu>2\delta \|W\|_{op} \,,
\eeqn 
and in particular, for all $j=1,\dots,Q$,
\eqn\label{eq-sig-DeltX0j-b-0-2}
	\sigma(W\Delta X_{0,j} + B) = W\Delta X_{0,j} + B
\eeqn
with $B=\mu \, u_Q u_{N_j}^T$, and any $\mu>2\delta \|W\|_{op}$.
\end{lemma}

\prf
We have 
\eqn
	Wx \in B_{\delta\|W\|_{op}}(0) \;\;\forall x\in B_\delta(0)\,,
\eeqn
and $B_{\delta\|W\|_{op}}(0)+b=B_{\delta\|W\|_{op}}(b)$. Given $b=\mu u_Q\in\R^Q$ with $\mu>2\delta\|W\|_{op}$, all points $x\in B_{\delta\|W\|_{op}}(b)$ lie in the positive sector $\R_+^Q\subset\R^Q$. Therefore, $\sigma$ maps any $x\in B_\delta(b)$ to itself.

From the definition of $\delta$ in \eqref{eq-delta-def-1-0}, 
\eqn 
	\Delta x_{0,j,i} \in  B_\delta(0)
	\;\;,\;
	\forall j=1,\dots,Q
	\;\;,\;
	\forall i=1,\dots, N_j \,.
\eeqn 
Therefore, 
\eqn 
	\sigma(W\Delta x_{0,j,i}+b)= W\Delta x_{0,j,i}+b \; \in\R^Q
\eeqn 
for all $j,i$, given $b=\mu u_Q\in\R^Q$ with $\mu>2\delta\|W\|_{op}$. 

Hence, we obtain \eqref{eq-sig-DeltX0j-b-0-2}.
\endprf

Using Lemma \ref{lm-sigma-b-2-0}, we repeat the recursive argument in the proof of Theorem \ref{thm-cC-uppbd-2} establishing \eqref{eq-tauW1b1-X0j-1-0} in the base case, and \eqref{eq-tauW1b1-X0j-1-1} in the induction step, for the same choices of $\underline W^{(\ell)}_*$ and $\underline b^{(\ell)}_*[\underline\mu]$, but for the parameter regime 
\eqn 
	\mu_i \in(2\delta|u_Q|,\infty) \;\;,\;i=1,\dots,\ell\,.
\eeqn  
Thereby, we obtain that, for $\ell=1,\dots,L$,    
\eqn\label{eq-Xell-notrunc-1-0}
	\tau_{\underline W^{(\ell)}_*,\underline b^{(\ell)}_*[\underline\mu]}(X_0) 
	= X_0 \,.
\eeqn 
%This is because $b^{(\ell)}$ translates all of the sets $W^{(\ell)}B_\delta(\overline{x_{0,j}})$, for $j=1,\dots,Q$, into the positive sector $\R_+^Q\subset\R^Q$ when $\mu_\ell>2\delta|u_Q|$. 
%We note that this mechanism is also used for the proof of Theorem 3.2 in \cite{cheewa-1}.
In particular, for $L=Q$, and $\underline\mu \in(2\delta|u_Q|,\infty)^Q$, we arrive at
\eqn 
	\tau_{\underline W_*^{(L)},\underline b_*^{(L)}[\underline\mu]}(X_0) = X_0 \,.
\eeqn 
Using the notation
\eqn
	\Delta_{2,*}^{(L)} &=& 
	\Delta_{1,*}^{(L)} \cN^{-1} (\Delta_{1,*}^{(L)})^{T} 
	\nonumber\\
	\Delta_{1,*}^{(L)} &=&  
	(\overline{(\tau_{\underline{ W }_*^{(L)} ,\underline{b}_*^{(L)}} (X_0) )^{red}})^{-1} 
	\Delta (\tau_{\underline{ W }_*^{(L)} ,\underline{b}_*^{(L)}} (X_0) ) \,,
\eeqn 
we find that \eqref{eq-Delt12-tau-def-0-1} and \eqref{eq-Delt12-tau-def-0-2}, \eqref{eq-Delt12-tau-def-0-3} imply that
\eqn 
	\Delta_{1,*}^{(L)} = \Delta_1^{(0)}
\eeqn 
and 
\eqn
	\Delta_{2,*}^{(L)} = \Delta_{2}^{(0)} 
\eeqn 
for all $\underline\mu\in(2\delta|u_Q|,\infty)^Q$.

From \eqref{eq-Thm3.2-CostN-1-0} thus follows that 
\eqn\label{eq-Thm3.2-CostN-1-2} 
	\widetilde\CostN[\underline W^{(L)}_*,\underline b^{(L)}_*[\underline\mu]]
	&:=&
	\min_{W_{L+1},b_{L+1}}
	\|
	W_{L+1}X^{(L)}+B_{L+1}-Y^{ext}\|_{\cL^2_{\cN}} 
	\nonumber\\ 
	&=&
	\big\|
 	Y|\Delta_2^{(L)}|^{\frac12}\big(1+\Delta_2^{(L)}\big)^{-\frac12}   
 	\big\|_{\cL^2}
	\nonumber\\
	&=&
	\big\|
 	Y|\Delta_2^{(0)}|^{\frac12}\big(1+\Delta_2^{(0)}\big)^{-\frac12}   
 	\big\|_{\cL^2} 
	\,,
\eeqn 
for all $\underline\mu\in(2\delta,\infty)^Q$.
All arguments thus far hold equally if, for $\ell=1,\dots,L$, we change $b^{(\ell)}_*$ (for any fixed choice of $\underline\mu\in(2\delta|u_Q|,\infty)^Q$) by an infinitesimal translation
\eqn
 	b^{(\ell)}_*[\underline\mu] \rightarrow  
 	b ^{(\ell)}:=b^{(\ell)}_*+\Delta b^{(\ell)}
\eeqn 
and $W^{(\ell)}$ by composition with an infinitesimal transformation 
\eqn 
	W^{(\ell)}_* \rightarrow  
	W ^{(\ell)} := W^{(\ell)}_*  (\1_{Q\times Q}+ \Delta W^{(\ell)}) 
\eeqn 
where $|\Delta\widetilde b^{(\ell)} |<\epsilon$ and $\|\Delta W^{(\ell)}\|_{op}<\epsilon$, for $\epsilon\ll1$ small enough.  
For these weights and biases, the least square minimization in \eqref{eq-Thm3.2-CostN-1-2} yields the same result, 
\eqn\label{eq-tildCostN-eps-1-0}
	\widetilde\CostN[\underline W^{(L)},\underline b^{(L)}]
	= \big\|
 	Y|\Delta_2^{(0)}|^{\frac12}\big(1+\Delta_2^{(0)}\big)^{-\frac12}   
 	\big\|_{\cL^2}  \,,
\eeqn 
where the r.h.s. is independent of the weights and biases. Since \eqref{eq-tildCostN-eps-1-0} is defined for an open $\epsilon$-neighborhood of weights and biases, we may differentiate it with respect to the components of $\underline W^{(L)}$ and $\underline b^{(L)}$, thus obtaining zero. Therefore, \eqref{eq-tildCostN-eps-1-0} corresponds to a degenerate local minimum because it was obtained from least square minimization. A more detailed presentation of this argument is given in the proof of Theorem 3.2 in \cite{cheewa-1}.

Finally, we note that by assumption \eqref{eq-DeltX0-neq0-1-2}, $\Delta X_{0,j}\neq0$ for all $j=1,\dots,Q$. Therefore, \eqref{eq-Thm3.2-CostN-1-0} is strictly positive, and thus, not a global minimum.
\qed

$\;$\\
\noindent
{\bf Acknowledgments:} 
T.C. gratefully acknowledges support by the NSF through the grant DMS-2009800, and the RTG Grant DMS-1840314 - {\em Analysis of PDE}. P.M.E. was supported by NSF grant DMS-2009800 through T.C. We thank Adam Klivans for very helpful comments.
\\

\end{document}